%% file: main.tex
\documentclass[letterpaper, 10 pt, conference]{ieeeconf}  

\IEEEoverridecommandlockouts                              

\usepackage{amsfonts}	
\usepackage{amsmath}	
\usepackage{amssymb}    
\usepackage{siunitx}
\usepackage{xfrac}    
\usepackage{pifont}   
\usepackage{dsfont}   
\usepackage{mathtools}   

\usepackage{booktabs}
\usepackage{makecell}  
\usepackage[flushleft]{threeparttable}  
\usepackage{multirow}
\usepackage{rotating}
\usepackage{microtype}
\usepackage{algorithm}
\usepackage{algorithmicx}
\usepackage{algpseudocode}  

\usepackage[dvipsnames]{xcolor}    
\let\labelindent\relax 
\definecolor{darkgreen}{rgb}{0.0,0.5,0.0}
\usepackage[inline]{enumitem} 
\usepackage{cite}  
\usepackage{xspace}    
\usepackage{tikz}
\usetikzlibrary{positioning,calc,arrows.meta}

\usepackage{array}
\usepackage{flushend}
\usepackage{float}

\renewcommand{\baselinestretch}{0.995}

\makeatletter
\let\NAT@parse\undefined
\makeatother

\usepackage[pdfencoding=auto, colorlinks=true]{hyperref} 
\usepackage[hang,flushmargin]{footmisc}

\usepackage[capitalize]{cleveref}
\crefname{section}{Sec.}{Secs.}
\Crefname{section}{Section}{Sections}
\Crefname{table}{Table}{Tables}
\crefname{table}{Tab.}{Tabs.}

\newcommand{\method}{\mbox{ParkDiffusion++}\xspace}

\title{\LARGE \bf
ParkDiffusion++: Ego Intention Conditioned Joint Multi-Agent Trajectory Prediction for Automated Parking using Diffusion Models}
\author{
Jiarong Wei$^{1,2}$,
Anna Rehr$^{2}$,
Christian Feist$^{2}$,
and Abhinav Valada$^{1}$
\thanks{$^1$ Department of Computer Science, University of Freiburg, Germany.}%
\thanks{$^2$ CARIAD SE, Germany.}%
\thanks{This work was partially funded by CARIAD SE.}
}

\begin{document}
\maketitle
\thispagestyle{empty}
\pagestyle{empty}


\begin{abstract}
    \input{sections/0_abstract}

\end{abstract}


\input{sections/1_introduction}
\input{sections/2_related_work}
\input{sections/3_method}
\input{sections/4_experiment}

\input{sections/5_conclusion}


\bibliographystyle{IEEEtran}
\bibliography{reference}


\end{document}

%% file: sections/0_abstract.tex
Automated parking is a challenging operational domain for advanced driver assistance systems, requiring robust scene understanding and interaction reasoning. The key challenge is twofold: (i)~predict multiple plausible ego intentions according to context and (ii)~for each intention, predict the joint responses of surrounding agents, enabling effective what‑if decision‑making.
However, existing methods often fall short, typically treating these interdependent problems in isolation.
We propose \method, which jointly learns a multi‑modal ego intention predictor and an ego conditioned multi-agent joint trajectory predictor for automated parking. Our approach makes several key contributions. First, we introduce an ego intention tokenizer that predicts a small set of discrete endpoint intentions from agent histories and vectorized map polylines. Second, we perform ego intention conditioned joint prediction, yielding socially consistent predictions of the surrounding agents for each possible ego intention. Third, we employ a lightweight safety‑guided denoiser with different constraints to refine joint scenes during training, thus improving accuracy and safety. Fourth, we propose counterfactual knowledge distillation, where an EMA teacher refined by a frozen safety‑guided denoiser provides pseudo‑targets that capture how agents react to alternative ego intentions. Extensive evaluations demonstrate that \method achieves state-of-the-art performance on the Dragon Lake Parking (DLP) dataset and the Intersections Drone (inD) dataset. Importantly, qualitative what‑if visualizations show that other agents react appropriately to different ego intentions.

%% file: sections/1_introduction.tex
\section{Introduction}

Automated parking systems are essential for managing shared spaces between human drivers and automated vehicles. 
A typical automated parking pipeline consists of three stages: searching, decision making, and maneuvering. 
The core challenge of decision making lies in coupling ego vehicle intention prediction with multi-agent joint what-if trajectory prediction conditioned on ego intention. 
Many existing trajectory prediction approaches operate in an open loop~\cite{distelzweig2025stochasticity, marvi2025evidential, distelzweig2025motion}, where future trajectories are predicted from past agent motions. 
However, they rarely model how surrounding agents would respond conditioned on counterfactual ego intentions.
This shortcoming is critical in parking lots, where weak lane semantics and mixed traffic induce diverse interactions including following a leading vehicle, yielding to oncoming traffic, and avoiding pedestrians. 
In addition, existing parking prediction works~\cite{shen2022parkpredict+, wei2025parkdiffusion, radwan2020multimodal} primarily yield marginal predictions without explicit scene‑level compatibility or safety constraints.


\begin{figure}[t]
    \centering\includegraphics[width=0.85\linewidth]{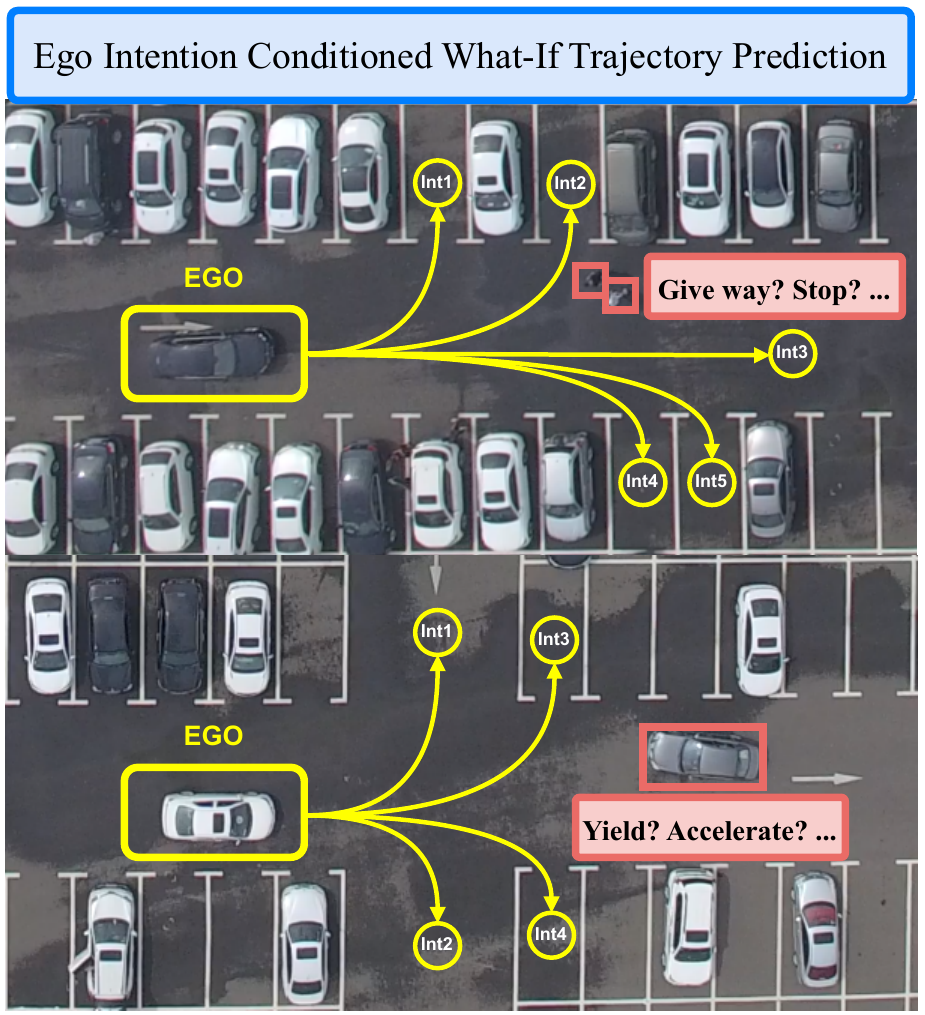}
    \caption{Parking lot environments, characterized by their unstructured nature relative to on-road driving, require complex reasoning for autonomous navigation. The ego vehicle must perform what-if prediction, a process of evaluating multiple feasible ego future intentions while conditionally predicting the reactive behaviors of other agents to ensure safe and efficient decision making. 
    }
    \label{fig:cover}
    \vspace*{-0.3cm}
\end{figure}

The task of combining ego intention prediction with ego intention conditioned joint trajectory prediction is challenging in two aspects.
First, real-world datasets provide ground truth only for the realized ego intention. Counterfactual intentions lack supervision, so training relies on reliable pseudo‑targets to cover unobserved what‑if queries.
Second, the introduction of intention selection and conditional decoding must generate scene-level trajectories that achieve high accuracy, while also ensuring safety and consistency.

In this work, we propose \method, which performs ego intention conditioned multi-agent joint trajectory prediction for automated parking, addressing both challenges. 
\method comprises two stages. 
Stage~1 trains an ego intention tokenizer that predicts a discrete set of ego intentions from local context features.
Stage~2 trains an ego intention conditioned joint predictor using ground truth and counterfactual intentions from Stage~1. 
Specifically, we build joint prediction on per‑agent marginals combined via a searching and selecting process. 
Then, an EMA teacher decoder refined by a safety‑guided denoiser provides pseudo targets for unobserved intentions.
We perform extensive evaluations on the Dragon Lake Parking and Intersections Drone datasets, where results show that \method sets the state-of-the-art for ego intention conditioned joint trajectory prediction in parking scenarios. 
We present detailed ablation studies and qualitative analysis of what-if predictions, demonstrating how other agents react to different ego intentions.

To summarize, our main contributions are as follows:

\begin{enumerate}
    \item The novel \method method for ego intention conditioned joint trajectory prediction. 
    \item A joint trajectory selector built on marginal models to ensure joint trajectory compatibility.
    \item A safety-guided denoiser to improve prediction safety. 
    \item A counterfactual knowledge distillation module conditioned on different ego intentions using the safety‑guided denoiser teacher.
    \item A new baseline for ego intention conditioned joint trajectory prediction in automated parking and state-of-the-art performance compared to existing methods. 
\end{enumerate} 

%% file: sections/2_related_work.tex
\section{Related Work}
In this section, we review the recent works that are related to and motivate our key contributions.

{\parskip=2pt
\noindent\textit{Joint Trajectory Prediction:}
Joint trajectory prediction targets a scene-consistent distribution over agent futures, rather than independent marginal forecasts. 
M2I~\cite{sun2022m2i} achieves joint trajectory prediction by identifying influencer–reactor agent pairs and combining marginal and conditional forecasts.
JFP~\cite{luo2023jfp} builds on MultiPath++~\cite{varadarajan2022multipath++}, modeling pairwise interactions with a graphical formulation to produce scene-consistent trajectories.
BiFF~\cite{zhu2023biff} fuses high-level future intentions with low-level interactive behaviors to yield joint, scene-consistent forecasts.
OptTrajDiff~\cite{wang2024optimizing} performs joint trajectory prediction using an optimized Gaussian diffusion process with clean-manifold guidance.
Trajectory Mamba~\cite{huang2025trajectory} replaces quadratic self-attention with a Selective-SSM encoder–decoder and joint polyline encoding to improve efficiency. 
Despite these advances, most joint predictors treat ego intent as latent and therefore cannot perform what-if trajectory prediction.
}

{\parskip=2pt
\noindent\textit{Ego Conditioned Trajectory Prediction:}
Ego conditioned prediction models how other agents respond when an explicit ego intention or trajectory is provided.
PiP~\cite{song2020pip} conditions multi-agent forecasting on candidate ego plans produced by a planner, coupling prediction with planning.
CBP~\cite{tolstaya2021identifying} predicts other agents’ future trajectories conditioned on a queried ego trajectory and quantifies interactivity.
WIMP~\cite{khandelwal2020wimp} injects counterfactual geometric goals or lanes to produce what-if forecasts of multi-actor reactions.
SceneTransformer~\cite{ngiam2022scenetransformer} unifies marginal, joint, and conditional modes and supports AV goal-conditioned prediction where other agents adapt to the chosen AV goal.
ScePT~\cite{chen2022scept} is a scene-consistent, policy-based joint predictor that supports conditional prediction by clamping one agent’s (e.g., the ego’s) future and rolling out coherent trajectories for the rest.
MotionLM~\cite{seff2023motionlm} casts multi-agent forecasting as language modeling over discrete motion tokens and supports temporally causal conditional rollouts for what-if queries.
DTPP~\cite{huang2024dtpp} employs a query-centric Transformer to perform efficient ego conditioned prediction for each branch in an ego trajectory tree.
Most recent methods assume an externally specified ego trajectory or goal at test time or rely on planner-generated branches.
To the best of our knowledge, none jointly learns a multi‑modal ego intention predictor and yields ego conditioned joint scene predictions.
}

{\parskip=2pt
\noindent\textit{Counterfactual Knowledge Distillation:} 
Peysakhovich~\textit{et al.}~\cite{peysakhovich2019robust} propose robust multi-agent counterfactual prediction from logged data, estimating outcomes under altered incentives or rules.
Hart and Knoll~\cite{hart2020counterfactual} evaluate autonomous-driving policies by constructing counterfactual worlds to assess agent reactions without extra labels.
Monti~\textit{et~al.}~\cite{monti2022many} distill a long-observation teacher into a short-horizon student to enable accurate, low-latency trajectory forecasts.
Bender~\textit{et~al.}~\cite{bender2023towards} distill causal behavior by generating counterfactual variants to correct Clever-Hans predictors.
Jung~\textit{et~al.}~\cite{jung2024counterfactually} analyze counterfactual fairness and align classifiers to fairness objectives via counterfactual knowledge distillation.
To the best of our knowledge, counterfactual knowledge distillation has not been applied to ego intention conditioned “what-if” trajectory prediction.
}

{\parskip=2pt
\noindent\textit{Guided Diffusion Models for Trajectory Prediction:}
Guided diffusion steers the reverse denoising process with auxiliary signals to generate intention or safety-aware motion trajectories.
MotionDiffuser~\cite{jiang2023motiondiffuser} applies constrained sampling with differentiable costs to produce rule-compliant, physically plausible multi-agent futures.
Diffusion-ES~\cite{yang2024diffusion} employs gradient-free evolutionary search with black-box rewards to guide diffusion-based trajectory generation.
Zheng~\textit{et~al.}~\cite{zheng2025diffusion} integrate classifier guidance into trajectory scores for a transformer-based diffusion planner, enabling safe, multi-modal closed-loop driving.
Plainer~\textit{et al.}~\cite{plainer2025consistent} introduce an energy-based diffusion model with a Fokker-Planck derived consistency regularizer in physical simulation, conceptually related to consistency-aware guidance. 
Inspired by these approaches, we incorporate safety-guidance components into our denoising process to more effectively enforce safety constraints and enhance model performance.
}

{\parskip=2pt
\noindent\textit{Trajectory Prediction for Automated Parking:}
ParkPredict~\cite{shen2020parkpredict} pioneered parking prediction, analyzing how model complexity and feature sets affect intent estimation and long-term forecasts in CARLA.
ParkPredict+~\cite{shen2022parkpredict+} addresses multimodal intent and trajectory prediction by combining CNNs and Transformers on a real-world parking dataset.
ParkDiffusion~\cite{wei2025parkdiffusion} accurately predicts the multimodal trajectories of vehicles and pedestrians using diffusion models. 
However, prior parking work predicts trajectories marginally, without enforcing scene compatibility or interactivity. Furthermore, these methods do not model how surrounding agents adapt to changes in ego intention, nor do they explicitly incorporate safety constraints.
}

%% file: sections/3_method.tex
\begin{figure*}[t]
    \centering
    \includegraphics[width=\textwidth]{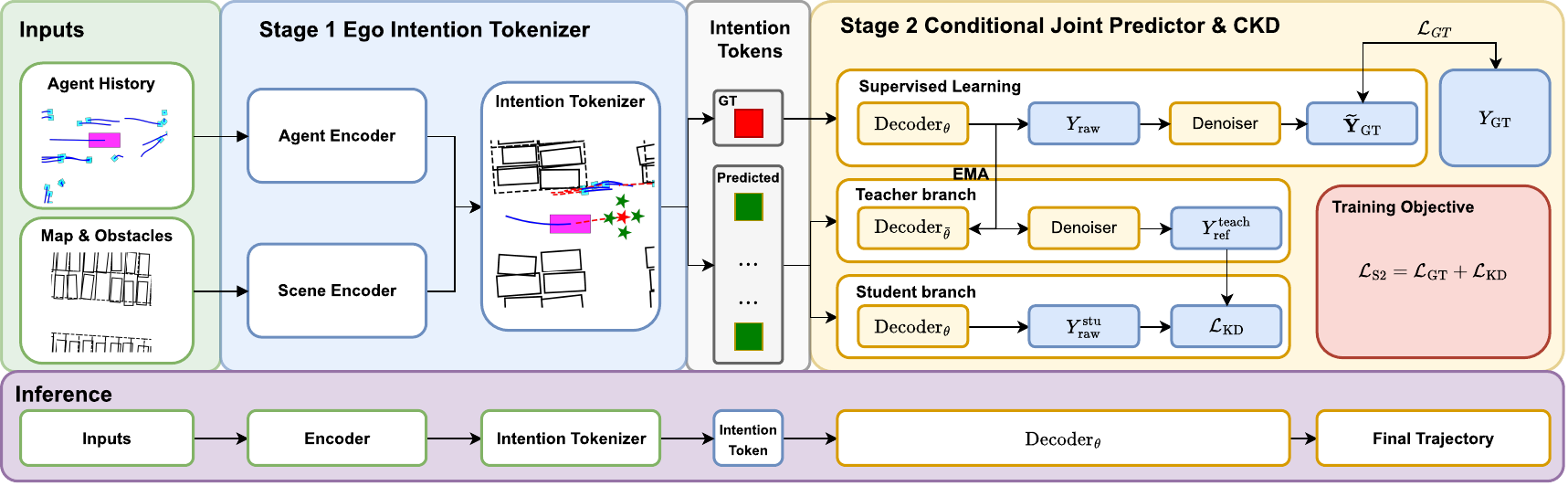}
    \vspace*{-0.6cm}
    \caption{\textbf{Overview of ParkDiffusion++.}
    Given agent histories and a vectorized map with parking slots and obstacles, Stage~1 predicts a bank of ego endpoint tokens, with the ground truth token in red and others in green.
    During Stage~2 training, we denote $\mathrm{Decoder}_\theta$ as the student conditional joint predictor, and $\mathrm{Decoder}_{\bar{\theta}}$ as EMA-updated teacher of $\mathrm{Decoder}_\theta$.
    In the supervised learning branch, the ground truth token drives $\mathrm{Decoder}_\theta$ to yield a raw joint scene. We compute $\mathcal{L}_{\text{GT}}$ on the raw output and use the frozen Denoiser to form a stop‑gradient consistency target (not shown for simplicity).
    The Counterfactual Knowledge Distillation (CKD) module consists of a teacher and a student branch.
    The teacher branch runs $\mathrm{Decoder}_{\bar{\theta}}$ and refines its output with the same frozen Denoiser as supervised learning to form a teacher target, and the student branch runs $\mathrm{Decoder}_\theta$ and learns to match that teacher target featuring an additional safety penalty.
    The primary training objective of Stage~2 comprises the supervised learning loss and the counterfactual knowledge distillation loss ($\mathcal{L}_{\text{KD}}$).
    At inference (bottom row), a token from Stage~1 is selected and fed to $\mathrm{Decoder}_\theta$ to yield the final joint trajectory.
    Internal steps such as per-agent marginals, beam-based assembly of joint scenes, and scene selection are abstracted within this block for clarity.}
    \label{fig:overview}
    \vspace*{-0.3cm}
\end{figure*}

\section{Method}
\label{sec:method}

In this section, we present our method for ego intention conditioned joint trajectory prediction, illustrated in Fig.~\ref{fig:overview}. 
We begin by formally defining the problem, then detail the components of our approach: an ego intention tokenizer, a conditional joint predictor featuring a reactive modeling and scoring mechanism, a pretrained safety-guided denoiser, and a Counterfactual Knowledge Distillation (CKD) scheme. The CKD queries alternative ego tokens and uses an Exponential Moving Average (EMA) teacher decoder refined by a frozen denoiser to produce safe pseudo‑targets.
Finally, we describe the two-stage training procedure that integrates these components.\looseness=-1

\subsection{Problem Formulation}
\label{subsec:formulation}

Given a scene observed in an ego-centric frame, the inputs consist of the past trajectories of $N$ agents, $\mathcal{X} = \{\mathbf{x}_{i,t}\}_{i=0}^{N-1}{}_{t=1}^{T_p}$, where $\mathbf{x}_{i,t} \in \mathbb{R}^{D}$ is the state of agent $i$ (ego is $i=0$) containing its position, heading, velocity, and acceleration at time $t$.
The inputs are augmented with scene context, $\mathcal{C}_{\mathrm{scene}}$, composed of agent attributes (e.g., category $c_i$, size $\mathbf{s}_i$) and a vectorized map representation describing parking slots and static obstacles.
Our core objective is two-fold. 
First, we aim to infer a probability distribution over a discrete set of $K_{\text{intent}}$ high‑level future intentions for the ego vehicle, $\mathcal{Z} = \{z_k\}_{k=1}^{K_{\text{intent}}}$, based on the observed past.
Second, and centrally, we aim to model a conditional distribution over the joint future trajectories of all agents, $\mathbf{Y} = \{\mathbf{y}_{i,t}\}_{i=0}^{N-1}{}_{t=1}^{T_f}$, given a specific ego intention $z_k$:
\begin{equation}
p_\theta\!\left(\mathbf{Y}\mid \mathcal{X}, \mathcal{C}_{\mathrm{scene}}, z_k\right).
\label{eq:main_prob}
\end{equation}

\subsection{Ego Intention Prediction}
\label{subsec:eif}
The first stage of our method is an intention tokenizer that proposes a discrete set of plausible ego endpoints. 
By representing future possibilities as a compact set of intention points rather than complete trajectories, we create a tractable and interpretable interface for downstream planning modules. This can then perform efficient what-if reasoning over a finite set of outcomes. \looseness=-1

\subsubsection{Scene Encoding}
\label{sec:scene_encoding}
The scene encoding architecture builds upon ParkDiffusion~\cite{wei2025parkdiffusion} to learn a comprehensive scene representation. 
Each agent's history $\mathbf{x}_{i,1:T_p}$ is processed by a temporal 1-D convolution followed by a GRU. The resulting sequence representation is then passed to a Transformer encoder over agents to capture local interactions, yielding $\mathbf{f}_i \in \mathbb{R}^d$. 
These features are then modulated by agent type via a learned affine transform to yield type-aware features $\tilde{\mathbf{f}}_i$. 
The vectorized map is processed in two streams to distinguish between soft constraints (drivable boundaries $\mathcal{M}_{\text{soft}}$) and hard constraints (physical obstacle segments $\mathcal{M}_{\text{hard}}$). 
The ego feature $\tilde{\mathbf{f}}_0$ queries each map stream via cross-attention. A global map summary $\mathbf{m}$ is obtained by max-pooling over these embedded map features.
Finally, the ego-agent's feature, a max-pooled summary of features from all other agents $\mathbf{f}_{\text{social}}$, and the global map summary are fused to obtain a holistic scene context vector $\mathbf{c}$.

\subsubsection{Discrete Ego Intention Tokenization}
We predict a discrete set of ego endpoints using a bank of learned mode embeddings $\{\mathbf{e}_k \in \mathbb{R}^{d_e}\}_{k=1}^{K_{\text{intent}}}$, which allows the model to discover a compact basis of common maneuvers directly from data. 
For each mode $k$, we compute $\mathbf{h}_k = f_{\mathrm{mode}}\!\big([\mathbf{c};\mathbf{e}_k]\big)$; linear heads regress the endpoint token and a logit \(s_k\), and a softmax over \(\{s_k\}_{k=1}^{K_{\text{intent}}}\) yields the categorical distribution.\looseness=-1

\subsection{Conditional Joint Prediction}
The second stage of our model is the conditional predictor, which generates full joint trajectories for all agents, contingent on a single ego intention token from the first stage. 

\subsubsection{Ego Conditioned Context}
The selected ego intention token (denoted as $\widehat{\mathbf{g}}_{k}$) is first processed by an MLP to yield a rich embedding $\boldsymbol{\tau}_k \in \mathbb{R}^{d_\tau}$.
For each agent $i$ in the scene, we first form the context $\mathbf{h}_i=[\tilde{\mathbf{f}}_i;\mathbf{f}_\text{social};\mathbf{m}] \in \mathbb{R}^{d_h}$ as described in Sec.~\ref{sec:scene_encoding}.
From \(\boldsymbol{\tau}_k\), we apply feature‑wise linear modulation (FiLM) to \(\mathbf{h}_i\), yielding the intention conditioned feature vector \(\mathbf{u}_{i,k}\).

\subsubsection{Reactive Social Modeling}
Our token-conditioned exposure gate learns to identify which agents are most relevant to the ego's intention, preventing over-sensitivity to irrelevant agents and thus preserving joint prediction accuracy.
Before computing the exposure, we define the terms in our formulation: $\sigma(\cdot)$ is the sigmoid function, $[\cdot]_+ = \max(0, \cdot)$ is the ReLU function, $d_{\mathrm{line}}(\cdot, \ell_k)$ is the minimum Euclidean distance from a point to the line segment $\ell_k$, and $\alpha, \beta, R_{\mathrm{path}}, R_{\mathrm{end}}$ are learnable scalar parameters.
Let $\ell_k$ be the line segment from the ego's last position to the intention token $\widehat{\mathbf{g}}_{k}$. We compute an exposure scalar for agent $i$, $e_i \in [0,1]$, based on its last observed position $\mathbf{p}^{\mathrm{last}}_i$:
\begin{align}
e_i &=\sigma\!\big(\alpha[R_{\mathrm{path}}-d_{\mathrm{line}}(\mathbf{p}^{\mathrm{last}}_i,\ell_k)]_+ \\\nonumber &+ \beta[R_{\mathrm{end}}-\|\mathbf{p}^{\mathrm{last}}_i-\widehat{\mathbf{g}}_{k}\|_2]_+\big).
\end{align}
A small gating network then uses this scalar to modulate the agent's feature vector $\mathbf{u}_{i,k}$.
This gate effectively learns a continuous, soft attention weight for each agent, conditioned on the ego's specific intention. 
Unlike hard attention, which would select a binary subset of agents, this soft mechanism allows the model to fluidly modulate the degree of influence for all agents based on their geometric relevance to the ego's intended maneuver.

\subsubsection{Joint Scene Prediction via Selection}
While we can decode $M$ plausible marginal futures for each of the $N$ agents, combining them into coherent joint scenes presents a combinatorial challenge. 
A brute-force approach would require evaluating $M^N$ possible scenes, which is computationally intractable for even modest numbers of agents and modes. 
To address this, we adopt a tractable three-stage pipeline. 
First, for each agent, we generate a set of $M$ marginal trajectory proposals along with compact per-mode features.
Second, we use a fast beam search over the top‑$R$ marginals per agent to construct a small, diverse set of \(K_{\text{scene}}\) high‑quality joint scene candidates. We use \(K_{\text{scene}}\) to denote the number of joint candidates.
Third, a learned scene selector scores and ranks these candidates. To evaluate a candidate scene, the selector scores each candidate using the global context and the sum of selected per-agent features. A softmax function is then applied to these scores for probability learning.
\vspace{-0.1cm}
\subsection{Safety-Guided Denoiser}
We develop a safety-guided denoiser, a powerful tool for refining trajectories by combining a learned data-driven prior with explicit rule-based guidance. 
The network $\boldsymbol{\epsilon}_\psi$ is framed within the paradigm of score-based generative diffusion modeling and is trained on a simple objective to predict the Gaussian noise $\boldsymbol{\xi}$ that was added to a clean trajectory $\mathbf{Y}_0$ at a given noise level $\sigma$. This process implicitly teaches the model the score function of the data distribution $\nabla_{\mathbf{Y}} \log p(\mathbf{Y})$, allowing it to understand what constitutes a "plausible" trajectory.

\subsubsection{Pretrained Denoiser}
We follow the structure of the Leapfrog Diffusion Model~\cite{mao2023leapfrog}, similar to ParkDiffusion, to pretrain the denoiser $\boldsymbol{\epsilon}_\psi$ independently and freeze it during Stage~2. 
When we use the denoiser to refine a trajectory during training, its output serves as a non-differentiable target, therefore no gradients flow into~$\psi$ in Stage~2. When used for refinement, the denoiser is conditioned on the same intention token as the decoder so that $C_{\text{tube}}$ and $C_{\text{end}}$ use the ego‑to‑token line segment $\ell_k$ and the token location $\mathbf{g}_k$.

\subsubsection{Deterministic Refinement with Geometric Guidance}
We denote this deterministic refinement operator by $\mathrm{Denoiser}_{\psi}(\cdot)$, 
which implements the project‑then‑guide steps using the pretrained score network $\boldsymbol{\epsilon}_\psi$.
Instead of a full, stochastic generation process, we use a few steps of a deterministic refinement algorithm called "project-then-guide." 
This process combines the learned data prior from the denoiser with an explicit guidance step from a differentiable potential function, $C(\mathbf{Y})$. 
The refinement at each step $s$ consists of two parts. 
First, in the projection step, we use the pretrained denoiser network to predict the noise in the current trajectory estimate $\mathbf{Y}^{(s)}$. 
Subtracting this noise yields an intermediate, cleaner trajectory, $\mathbf{Y}^{(s+1/2)}$, effectively projecting it onto the learned data manifold. 
Second, in the guidance step, we apply an explicit correction to this plausible trajectory using the gradient of our geometric potential function. 
This function acts as a differentiable critic, and its gradient provides a corrective force that pushes the trajectory away from constraint violations. 
The full update is:
\begin{align}
\begin{split}
\mathbf{Y}^{(s+1/2)} &= \mathbf{Y}^{(s)} - \boldsymbol{\epsilon}_\psi(\mathbf{Y}^{(s)},\sigma_s;\mathcal{C}_{\mathrm{scene}}, \mathbf{g}_k), \\
\mathbf{Y}^{(s+1)} &= \mathbf{Y}^{(s+1/2)} - \eta_s\,\nabla_{\mathbf{Y}} C(\mathbf{Y}^{(s+1/2)}).
\end{split}
\end{align}

\subsubsection{Geometric Potential Functions}
The potential function $C(\mathbf{Y})$ sums five key terms that enforce safety, intention consistency, and physical plausibility:
\begin{itemize}
    \item Agent-Agent Overlap ($C_{\mathrm{ov}}$): A hinge loss on the distance between agents' safety radii ($r_i, r_j$), denoted as
    \begin{equation}
    C_{\mathrm{ov}}=\sum_{t,i\neq j}\!\big[(r_i\!+\!r_j-\|\mathbf{y}_{i,t}\!-\!\mathbf{y}_{j,t}\|_2)_+\big]^2.
    \end{equation}
    \item Obstacle Clearance ($C_{\mathrm{obs}}$): A hinge loss penalizing proximity to static obstacle segments, defined as
    \begin{equation}
    C_{\mathrm{obs}}=\sum_{i,t}\big[(m_{\mathrm{obs}}-d^{\min}_{i,t})_+\big]^2,
    \end{equation}
    where $m_{\mathrm{obs}}$ and $d^{\min}_{i,t}$ denote obstacle clearance margin, and Euclidean distance from the agent $i$'s rectangular footprint at time $t$, respectively.
    \item Ego Path-Tube ($C_{\mathrm{tube}}$): Encourages ego trajectory $\mathbf{y}_{0,t}$ to stay within a radius $R_{\mathrm{tube}}$ of its intended path $\ell_k$.
    \item Ego Endpoint Anchoring ($C_{\mathrm{end}}$): Ensures ego trajectory terminates near the conditioning intention token $\widehat{\mathbf{g}}_k$.
    \item Motion Smoothness ($C_{\mathrm{sm}}$): Enforces kinematic plausibility by penalizing velocity and acceleration magnitudes.
\end{itemize}

\subsection{Counterfactual Knowledge Distillation}
\label{subsec:cfkd}
A key limitation of imitation learning is that datasets only provide supervision for a single reality that occurred. 
To enable our model to reason about what would have happened under different ego intentions, we propose a Counterfactual Knowledge Distillation (CKD) framework. 

Let $\mathrm{Decoder}_\theta$ denote the student decoder.
We maintain an exponential moving average (EMA) teacher \(\mathrm{Decoder}_{\bar{\theta}}\) of the student \(\mathrm{Decoder}_\theta\), updated after each step by
\begin{equation}
\bar{\theta} \leftarrow \tau\,\bar{\theta} + (1-\tau)\,\theta, \qquad \tau \in [0.99,\,0.999].
\label{eq:ema_update}
\end{equation}
The denoiser and tokenizer remain frozen in Stage~2.

We sample a counterfactual token \(\widehat{\mathbf{g}}_{\tilde{k}}\) from the Stage~1 bank (\(\tilde{k}\neq k^*\)), then form a teacher target by first predicting with the EMA teacher-decoder and then refining with the frozen denoiser:
\vspace{-0.1cm}
\begin{align}
\mathbf{Y}^{\text{teach}}_{\text{raw}} &= \mathrm{Decoder}_{\bar{\theta}}(\mathcal{X}, \mathcal{C}_{\mathrm{scene}}, \widehat{\mathbf{g}}_{\tilde{k}}),\\
\mathbf{Y}^{\text{teach}}_{\text{ref}} &= \mathrm{Denoiser}_{\psi}(\mathbf{Y}^{\text{teach}}_{\text{raw}}, \mathcal{C}_{\mathrm{scene}}, \widehat{\mathbf{g}}_{\tilde{k}}).
\end{align}
We detach this refined target $\operatorname{sg}[\mathbf{Y}^{\text{teach}}_{\text{ref}}]$ to prevent gradients flowing into the teacher. The \emph{student} prediction is produced by the current decoder without refinement:
\vspace{-0.1cm}
\begin{equation}
\mathbf{Y}^{\text{stu}}_{\text{raw}}=\mathrm{Decoder}_{\theta}(\mathcal{X}, \mathcal{C}_{\mathrm{scene}}, \widehat{\mathbf{g}}_{\tilde{k}}).
\end{equation}
The distillation loss encourages the student to approximate the teacher's refined scene while remaining collision-free:
\vspace{-0.1cm}
{\small
\begin{equation}
\mathcal{L}_{\mathrm{KD}} \,=\, \lambda_{\mathrm{kd}} \left\|\mathbf{Y}^{\text{stu}}_{\text{raw}} - \operatorname{sg}\!\big[\mathbf{Y}^{\text{teach}}_{\text{ref}}\big]\right\|_2^2
\;+\; \lambda_{\mathrm{safe}}\,\mathrm{CoL}_{\delta}\!\left(\mathbf{Y}^{\text{stu}}_{\text{raw}}\right),
\label{eq:lkd}
\end{equation}}
where \(\mathrm{CoL}_\delta(\cdot)\) denotes a differentiable collision‑overlap penalty with clearance \(\delta\).
Only $\theta$ is updated by $\mathcal{L}_{\mathrm{KD}}$, $\bar{\theta}$ is updated by EMA~\eqref{eq:ema_update} and $\psi$ is frozen.

\subsection{Training Scheme}
Our model is trained in two distinct stages to decouple intention discovery from intention conditioned dynamics.

\subsubsection{Stage~1: Ego Intention Tokenizer}
The objective is to accurately model the distribution of realized ego intentions.
We use a "winner-takes-all" loss to optimize SmoothL1 for endpoint regression, cross-entropy for mode selection, and a diversity regularizer, weighted by $\lambda_\text{xy}$, $\lambda_\text{cls}$, $\lambda_\text{div}$.

\subsubsection{Stage~2: Conditioned Joint Prediction}
The primary objective $\mathcal{L}_{\mathrm{SUP}}$ uses the ground truth ego intention for direct supervision on both the per-agent marginal predictions and the final selected joint scene. We train with two branches: a supervised (GT) branch and, with probability $p_{cf}$, an additional counterfactual KD branch.
Given the ground truth intention token denoted as $\widehat{\mathbf{g}}_{k^*}$, the student decoder predicts a raw joint scene
$\mathbf{Y}_{\text{raw}}=\mathrm{Decoder}_{\theta}(\mathcal{X}, \mathcal{C}_{\mathrm{scene}}, \widehat{\mathbf{g}}_{k^*})$,
which we refine with the frozen denoiser to obtain
$\widetilde{\mathbf{Y}}_{\text{GT}}=\mathrm{Denoiser}_{\psi}(\mathbf{Y}_{\text{raw}}, \mathcal{C}_{\mathrm{scene}}, \widehat{\mathbf{g}}_{k^*})$.
We supervise the raw prediction against ground truth and include a stop‑gradient consistency term that encourages cooperation with the frozen refiner:\looseness=-1
\begin{equation}
\begin{split}
\mathcal{L}_{\mathrm{GT}} \,=\, &\lambda_{\mathrm{raw}}\,\mathcal{L}_{\mathrm{SUP}}\!\left(\mathbf{Y}_{\text{raw}}, \mathbf{Y}^{\text{gt}}\right) \\
&+\, \lambda_{\mathrm{cons}}\,\left\|\mathbf{Y}_{\text{raw}} - \operatorname{sg}\!\big[\widetilde{\mathbf{Y}}_{\text{GT}}\big]\right\|_2^2 \\
&+\, \lambda_{\mathrm{safe}}\,\mathrm{CoL}_{\delta}\!\left(\mathbf{Y}_{\text{raw}}\right).
\end{split}
\label{eq:lgt}
\end{equation}
Gradients from $\mathcal{L}_{\mathrm{GT}}$ update $\theta$ only; $\psi$ remains frozen.

For each batch, we always run the supervised branch and, with probability $p_{cf}$, additionally run the counterfactual distillation branch. The Stage~2 objective is
\begin{equation}
\mathcal{L}_{\mathrm{S2}} \,=\, \mathcal{L}_{\mathrm{GT}} \;+\; \mathbb{I}\{\text{CF}\}\cdot \mathcal{L}_{\mathrm{KD}},
\end{equation}
where \(\text{CF}\sim\mathrm{Bernoulli}(p_{\mathrm{cf}})\) indicates whether the counterfactual branch is used.
Only the student parameters $\theta$ receive gradients, and the denoiser $\psi$ and tokenizer are frozen. The teacher-decoder parameters $\bar{\theta}$ are updated by the EMA rule in \eqref{eq:ema_update}.
This hybrid scheme ensures the model accurately predicts realized outcomes while learning to generate safe and plausible responses to a wide range of counterfactual plans.\looseness=-1

%% file: sections/4_experiment.tex
\section{Experiments}
\label{sec:exp}

This section first presents the datasets, metrics, and baselines that we use. 
We then detail the implementation, followed by quantitative and qualitative results and an ablation study on each component in our model.

\begin{table*}[t]
\centering
\caption{Benchmarking results on the Dragon Lake Parking (DLP) dataset.}
\vspace*{-0.3cm}
\label{table:results-dlp}
\setlength\tabcolsep{5pt}
\begin{threeparttable}
    \begin{tabular}{l | ccccc | ccccc}
        \toprule
        \textbf{Model}
        & \textbf{minADE}[\unit{\meter}] & \textbf{minFDE}[\unit{\meter}] & \textbf{minMR} & \textbf{minOR} & \textbf{mAP} 
        & \textbf{f-ADE}[\unit{\meter}] & \textbf{f-FDE}[\unit{\meter}] & \textbf{f-MR} & \textbf{f-OR} & \textbf{f-mAP} \\
        \midrule
        ParkDiffusion$^\dag$~\cite{wei2025parkdiffusion} & 0.52 & 1.06 & 0.48 & 0.21 & 0.68 & 0.90 & 1.78 & 0.63 & 0.28 & 0.57 \\
        WIMP~\cite{khandelwal2020wimp} & 0.62 & 1.20 & 0.46 & 0.18 & 0.70 & 0.84 & 1.60 & 0.55 & 0.24 & 0.62 \\
        SceneTransformer~\cite{ngiam2022scenetransformer}  & 0.36 & 0.79 & 0.28 & 0.07 & 0.92 & 0.55 & 1.08 & 0.41 & 0.12 & 0.84 \\
        ScePT~\cite{chen2022scept}  & 0.38 & 0.86 & 0.30 & 0.08 & 0.90 & 0.61 & 1.28 & 0.37 & 0.11 & 0.88 \\
        MotionLM~\cite{seff2023motionlm}    & \underline{0.32} & \underline{0.80} & 0.29 & 0.08 & \underline{0.95} & \textbf{0.34} & \underline{0.88} & 0.35 & 0.10 & \underline{0.94} \\
        DTPP~\cite{huang2024dtpp}  & \underline{0.32} & 0.82 & \underline{0.27} & \underline{0.06} & 0.94 & 0.50 & 1.21 & \underline{0.33} & \underline{0.09} & 0.92 \\
        \midrule
        ParkDiffusion++ (Ours) 
                          & \textbf{0.29} & \textbf{0.56} & \textbf{0.23} & \textbf{0.03} & \textbf{0.97} 
                          & \underline{0.36} & \textbf{0.66} & \textbf{0.29} & \textbf{0.05} & \textbf{0.95} \\
        \bottomrule
    \end{tabular}
    We report results for multiple baselines on the Dragon Lake Parking (DLP) dataset~\cite{shen2022parkpredict+} validation split.
    Evaluation metrics include oracle (min-) and final (f-) ADE, FDE, MR, OR, mAP.
    Lower is better for ADE/FDE/MR/OR, and higher is better for mAP. 
    Bold and underline denote best and second‑best results, respectively.
\end{threeparttable}
\end{table*}
\vspace{-0.2cm}
\begin{table*}[t]
\centering
\caption{Benchmarking results on the Intersections Drone (inD) dataset.}
\vspace*{-0.3cm}
\label{table:results-ind}
\setlength\tabcolsep{5pt}
\begin{threeparttable}
    \begin{tabular}{l | ccccc | ccccc}
        \toprule
        \textbf{Model}
        & \textbf{minADE}[\unit{\meter}] & \textbf{minFDE}[\unit{\meter}] & \textbf{minMR} & \textbf{minOR} & \textbf{mAP} 
        & \textbf{f-ADE}[\unit{\meter}] & \textbf{f-FDE}[\unit{\meter}] & \textbf{f-MR} & \textbf{f-OR} & \textbf{f-mAP} \\
        \midrule
        ParkDiffusion$^\dag$~\cite{wei2025parkdiffusion}& 0.86 & 1.60 & 0.58 & 0.31 & 0.66 & 1.12 & 2.05 & 0.68 & 0.38 & 0.54 \\
        WIMP~\cite{khandelwal2020wimp} & 0.78 & 1.48 & 0.52 & 0.28 & 0.72 & 0.99 & 1.82 & 0.61 & 0.33 & 0.60 \\
        SceneTransformer~\cite{ngiam2022scenetransformer} & 0.58 & 1.10 & 0.43 & 0.19 & 0.84 & 0.86 & 1.65 & 0.51 & 0.22 & 0.69 \\
        ScePT~\cite{chen2022scept} & \textbf{0.40} & \underline{0.88} & \underline{0.36} & 0.15 & \underline{0.91} & 0.67 & 1.35 & 0.47 & 0.18 & 0.74 \\
        MotionLM~\cite{seff2023motionlm} & 0.48 & 1.05 & 0.39 & 0.15 & 0.89 & \underline{0.63} & 1.31 & \underline{0.43} & 0.18 & \underline{0.81} \\
        DTPP~\cite{huang2024dtpp} & 0.49 & 0.96 & 0.38 & \underline{0.13} & 0.87 & \textbf{0.61} & \underline{1.22} & 0.45 & \underline{0.16} & 0.76 \\
        \midrule
        ParkDiffusion++ (Ours) 
                          & \underline{0.42} & \textbf{0.78} & \textbf{0.32} & \textbf{0.07} & \textbf{0.93} 
                          & 0.66 & \textbf{1.18} & \textbf{0.38} & \textbf{0.09} & \textbf{0.83} \\
        \bottomrule
    \end{tabular}
    We also report results on the inD (drone‑based intersections) dataset~\cite{bock2020ind} validation split. 
    Evaluation metrics include oracle (min-) and final (f-) ADE, FDE, MR, OR, mAP.
    Lower is better for ADE/FDE/MR/OR, and higher is better for mAP. 
    Bold and underline denote best and second‑best results, respectively.
\end{threeparttable}
\end{table*}

\begin{figure*}[t]
    \centering
    \begin{minipage}[b]{0.24\textwidth}
        \centering
        \includegraphics[width=\linewidth]{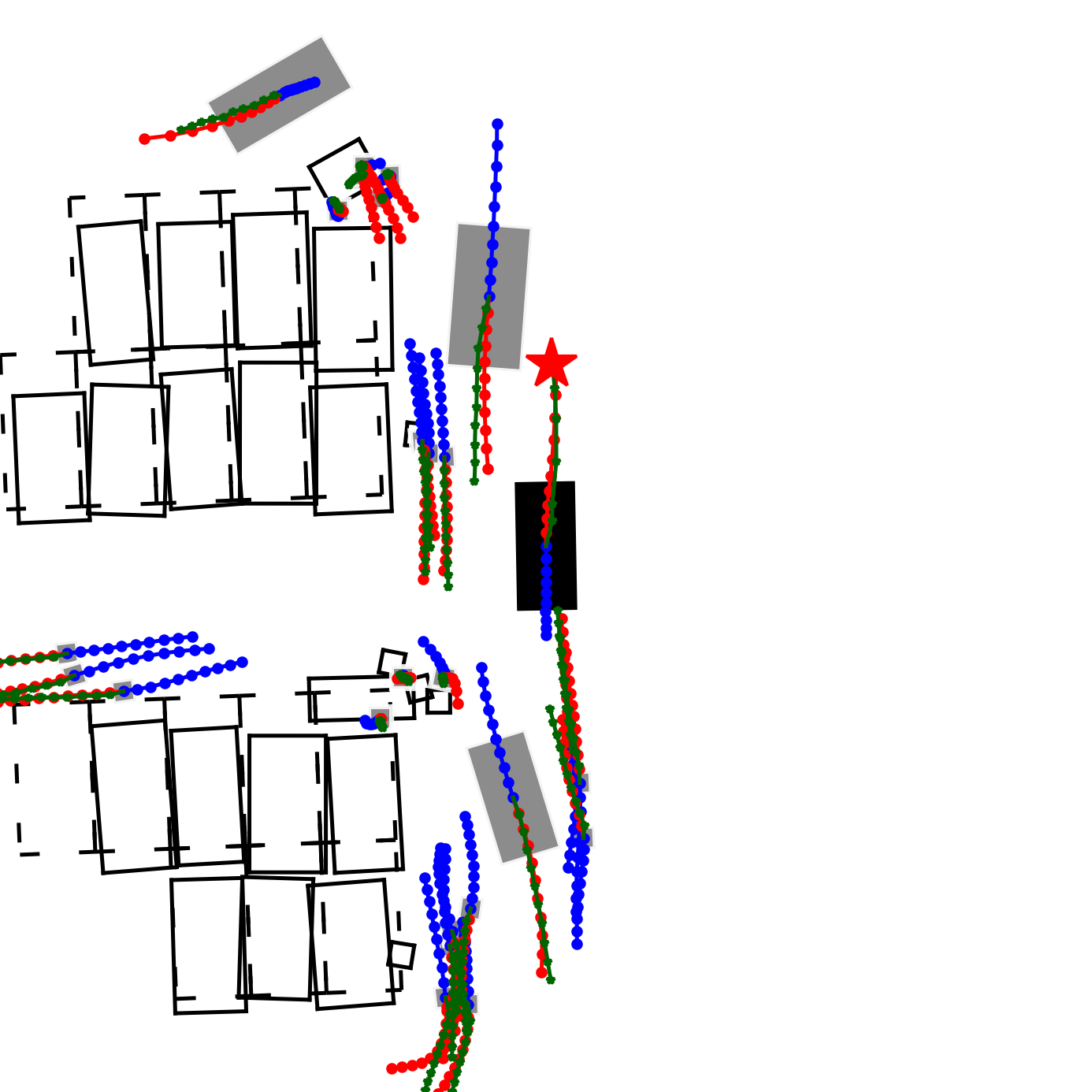}\\[0.3em]
    \end{minipage}
    \hfill
    \begin{minipage}[b]{0.24\textwidth}
        \centering
        \includegraphics[width=\linewidth]{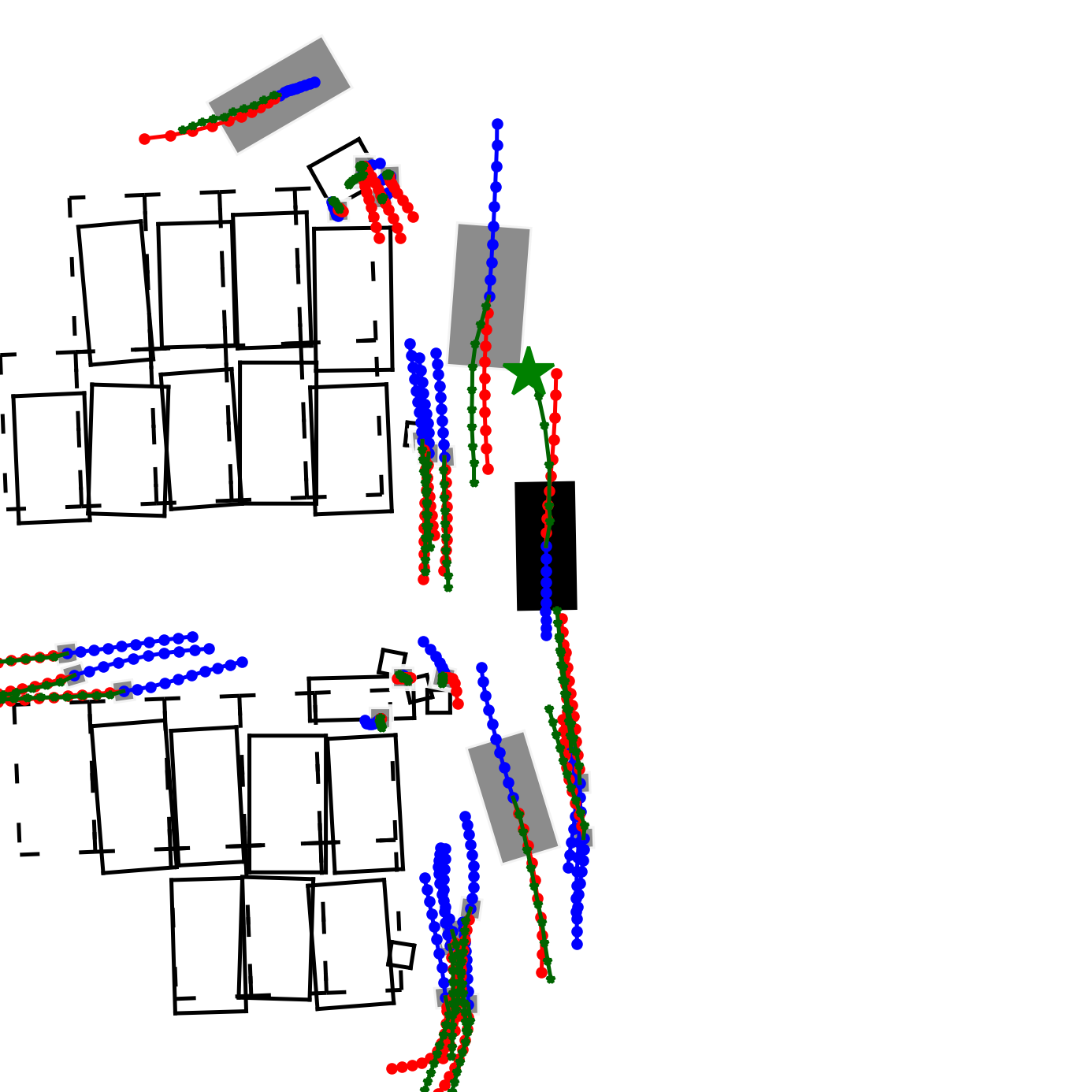}\\[0.3em]
    \end{minipage}
    \hfill
    \begin{minipage}[b]{0.24\textwidth}
        \centering
        \includegraphics[width=\linewidth]{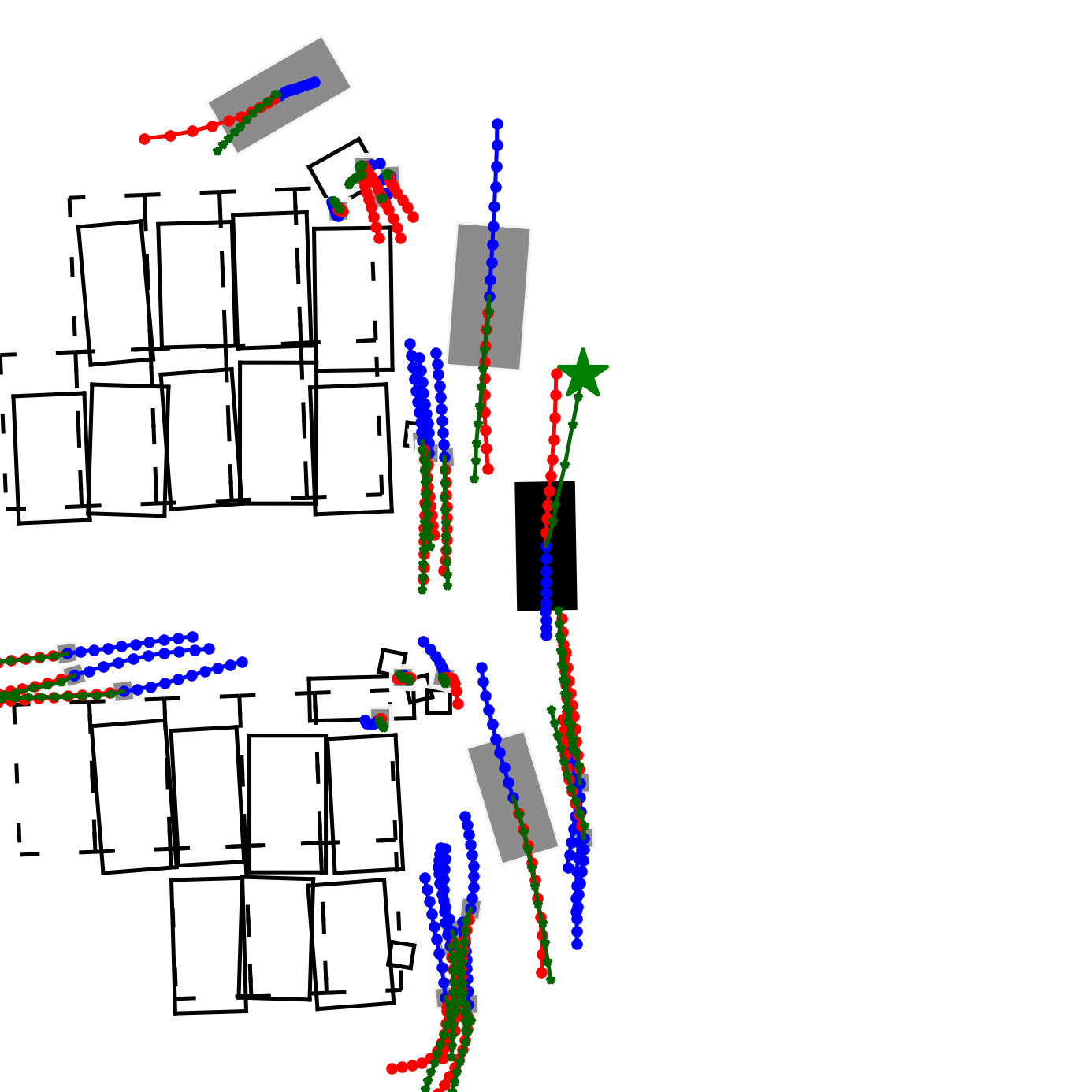}\\[0.3em]
    \end{minipage}
    \hfill
    \begin{minipage}[b]{0.24\textwidth}
        \centering
        \includegraphics[width=\linewidth]{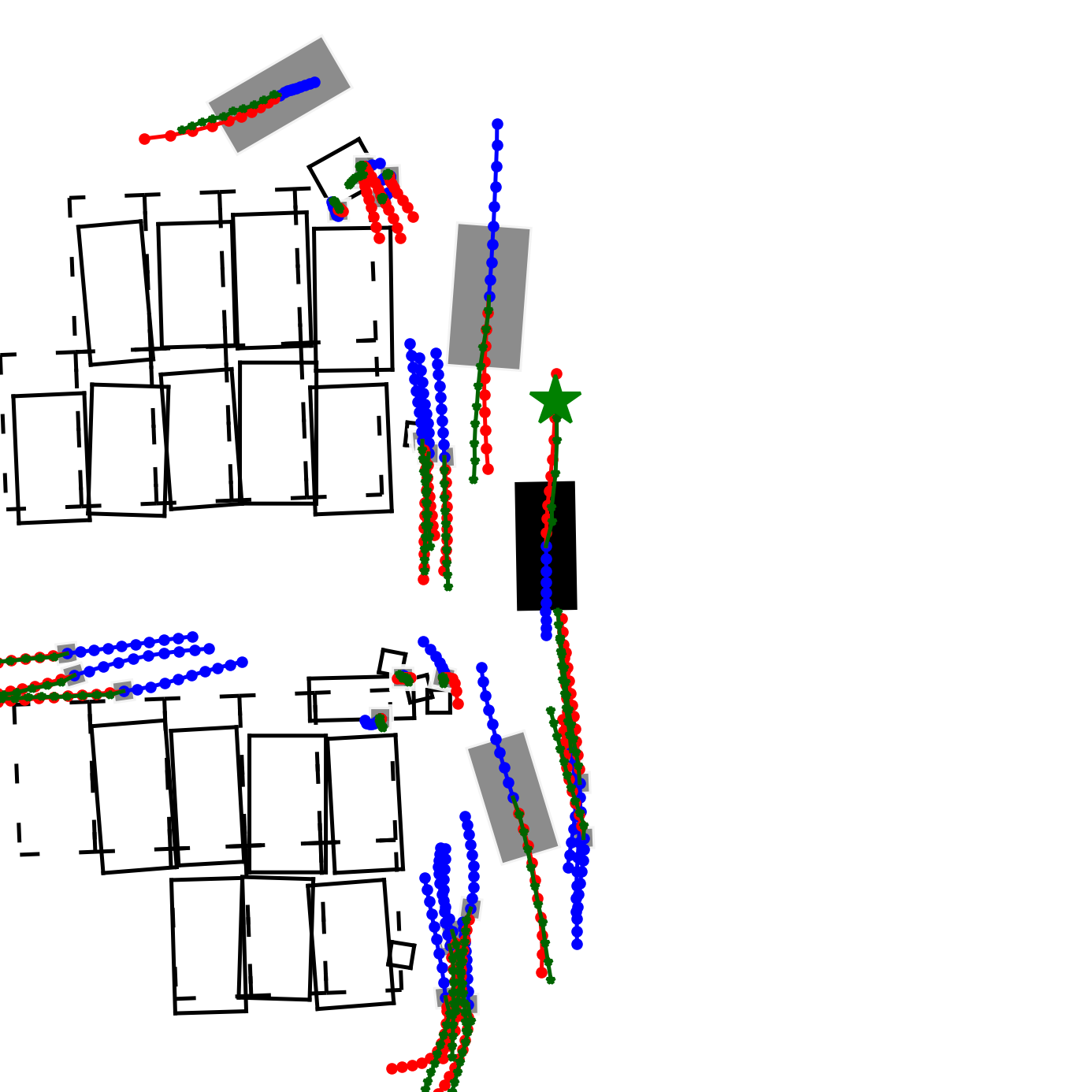}\\[0.3em]
    \end{minipage}
    
    \vspace{0.5cm}
    
    \begin{minipage}[b]{0.24\textwidth}
        \centering
        \includegraphics[width=\linewidth]{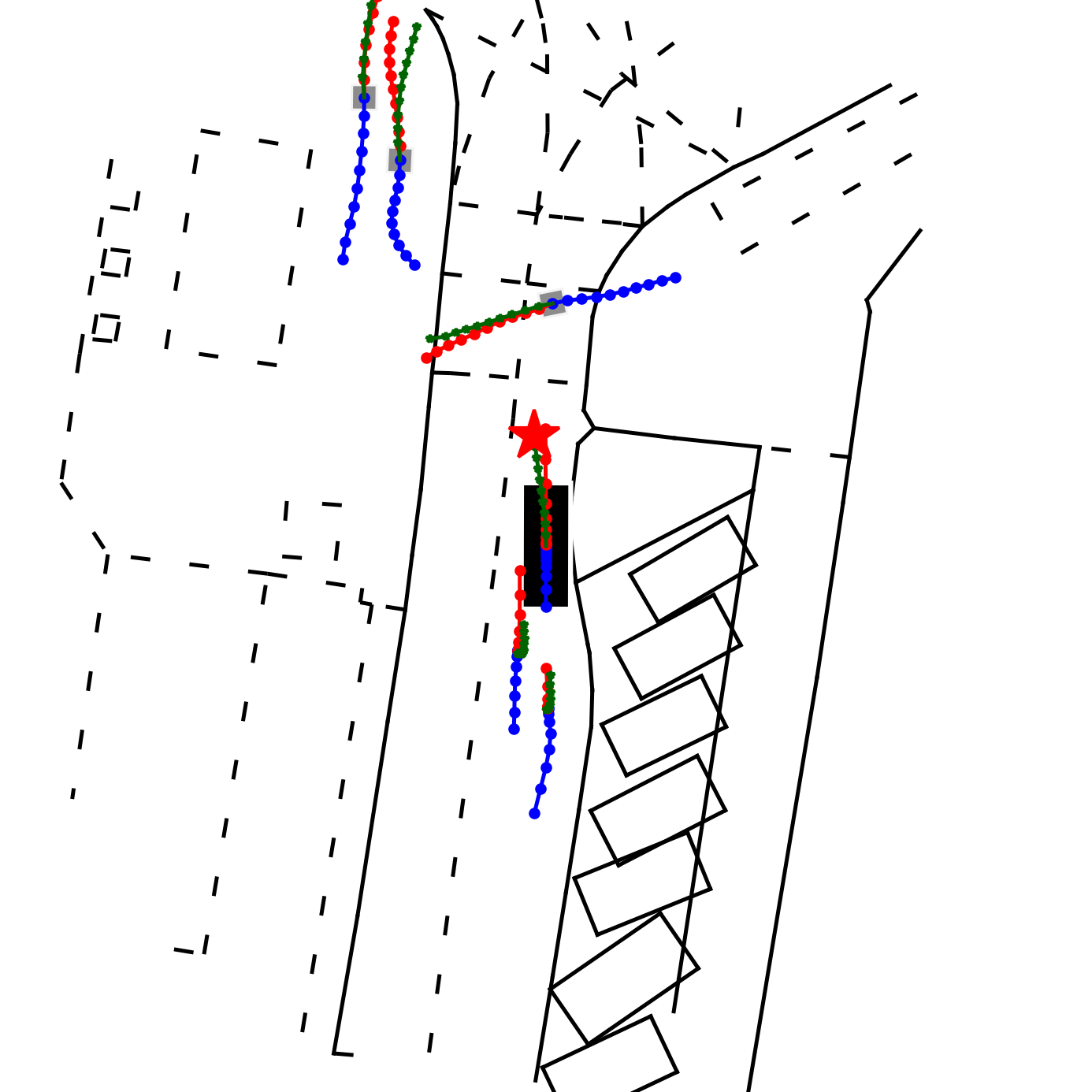}\\[0.3em]
    \end{minipage}
    \hfill
    \begin{minipage}[b]{0.24\textwidth}
        \centering
        \includegraphics[width=\linewidth]{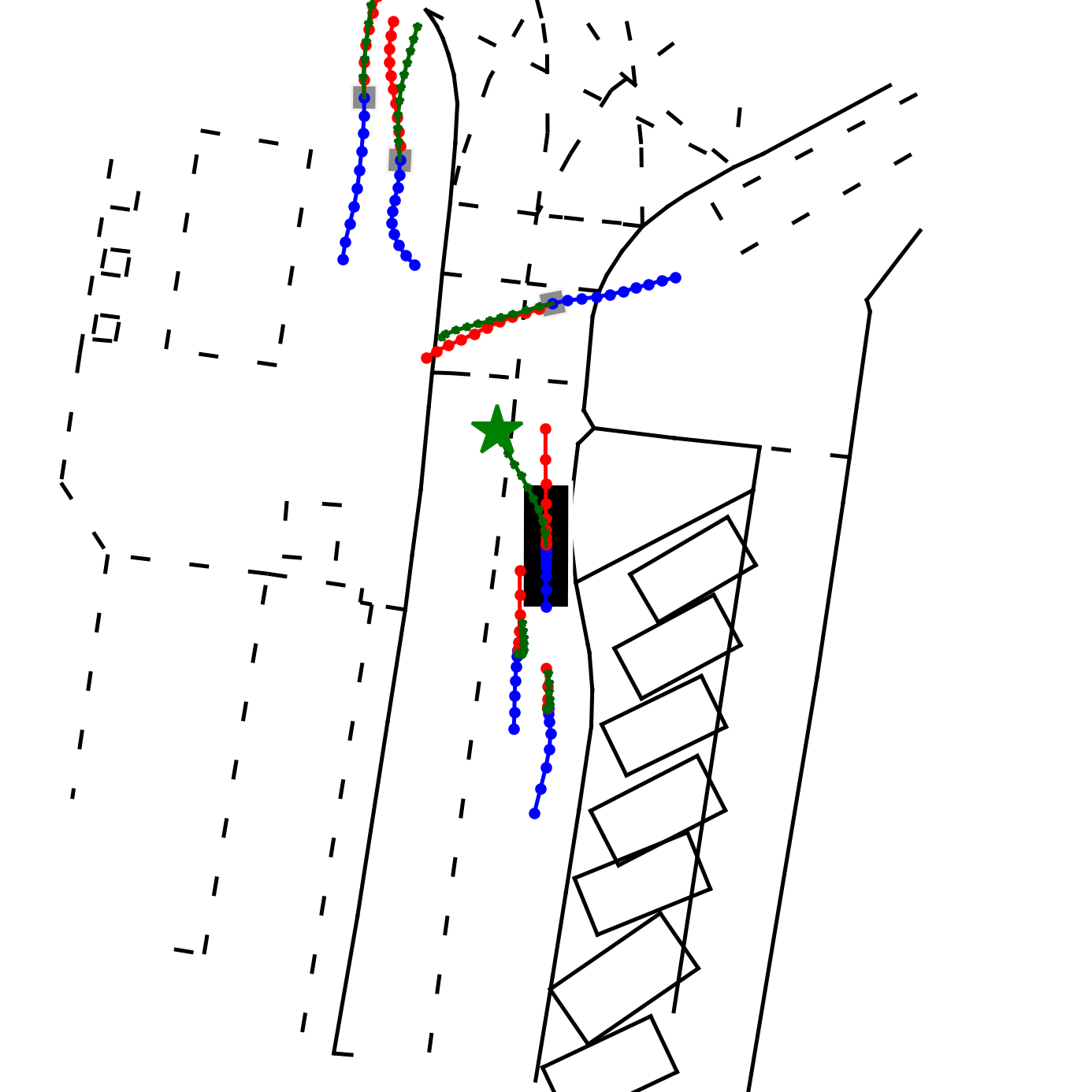}\\[0.3em]
    \end{minipage}
    \hfill
    \begin{minipage}[b]{0.24\textwidth}
        \centering
        \includegraphics[width=\linewidth]{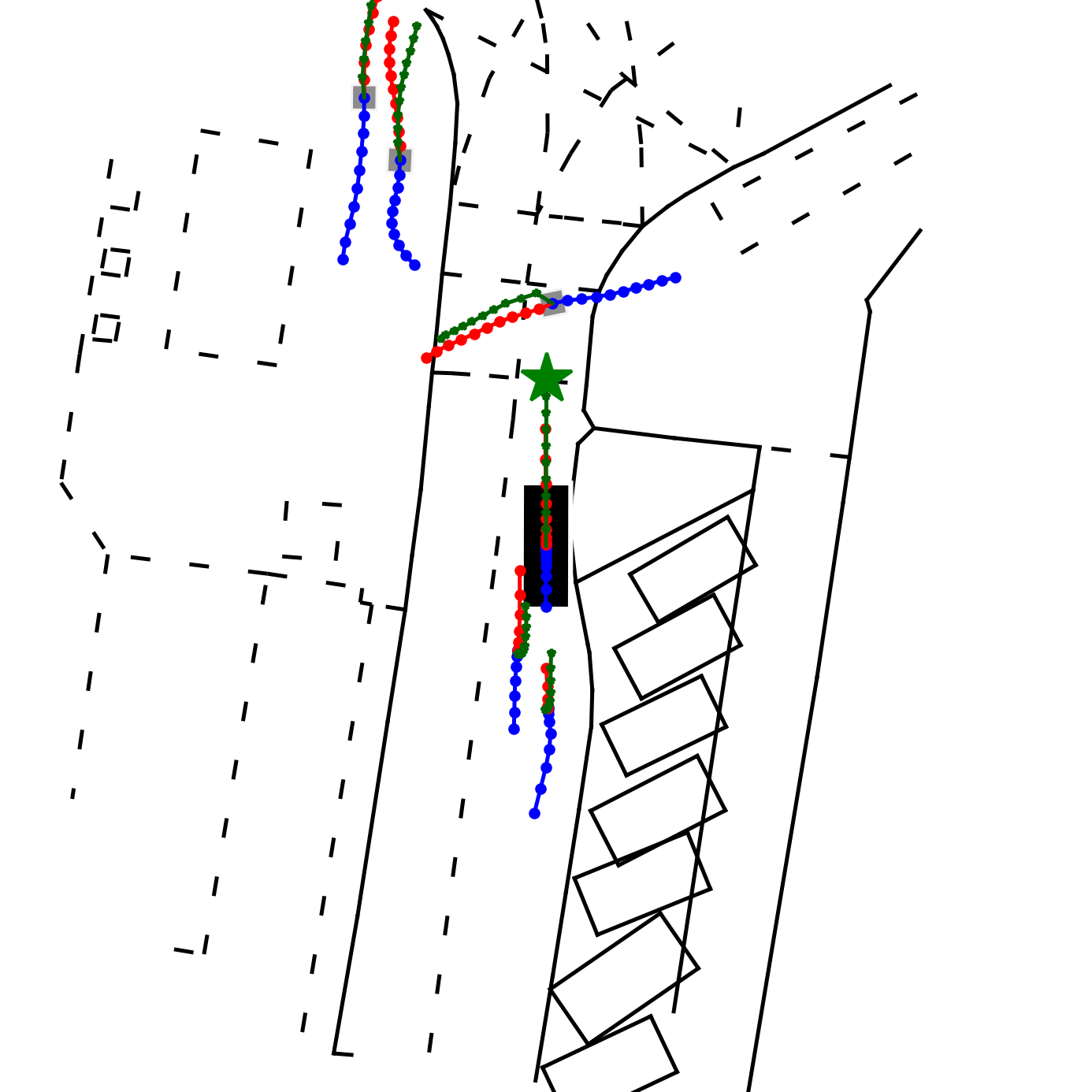}\\[0.3em]
    \end{minipage}
    \hfill
    \begin{minipage}[b]{0.24\textwidth}
        \centering
        \includegraphics[width=\linewidth]{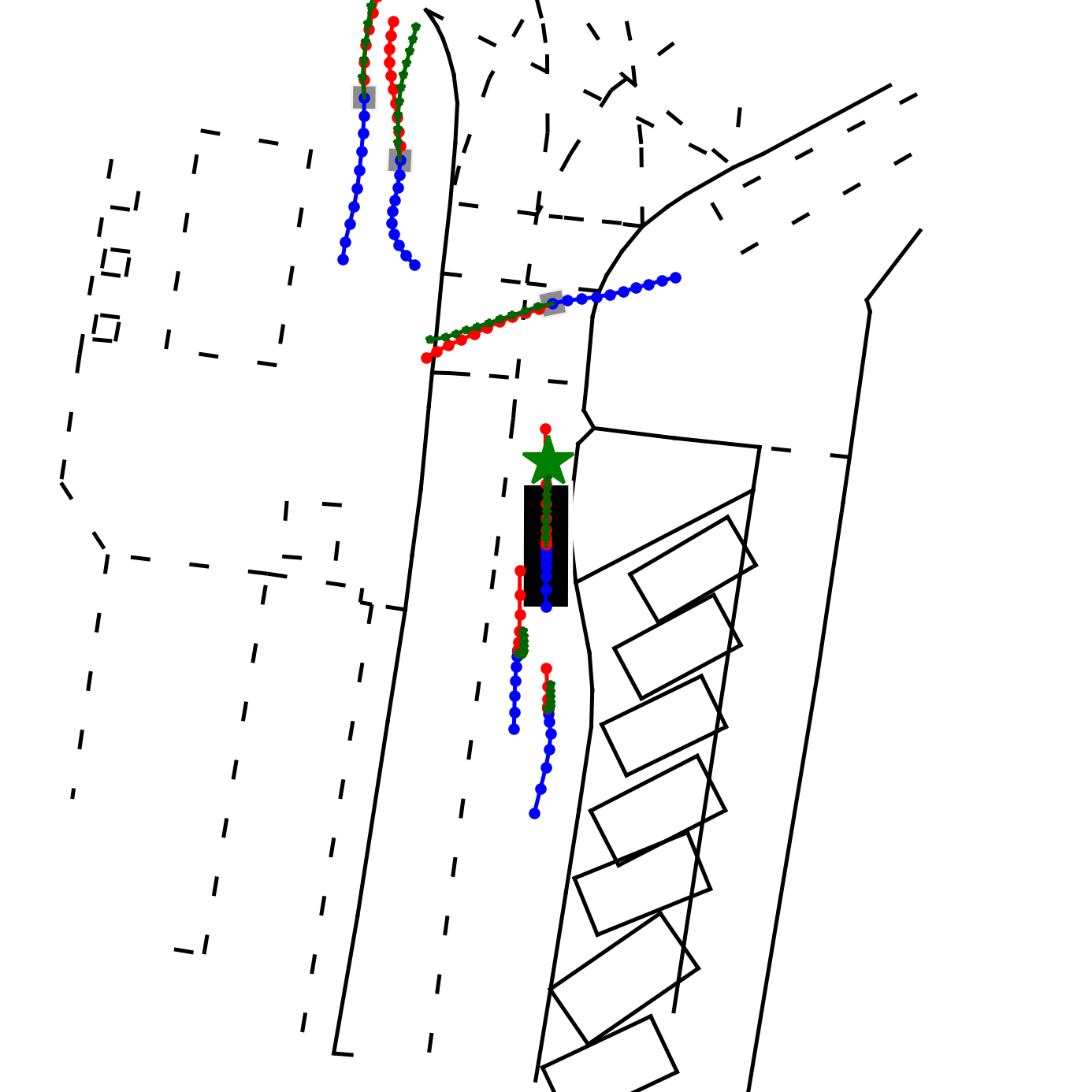}\\[0.3em]
    \end{minipage}
    
    \caption{Visualizations of \method predictions on the DLP (\textit{top row}) and inD (\textit{bottom row}) datasets. For both rows, the leftmost figure shows how our model reacts to the most likely intention (red star) from Stage~1. The other three figures in each row show the joint prediction result conditioned on other representative predicted intentions (green star). The ego vehicle is depicted in solid black at the center, while other agents are shown in gray. Trajectories are color-coded as follows: \textcolor{blue}{past} in blue, \textcolor{red}{ground-truth future} in red, and \textcolor{darkgreen}{predicted} in green.}
    \label{fig:main_figure}
    \vspace*{-.3cm}
\end{figure*}

\subsection{Experimental Setup}
\label{subsec:setup}

\subsubsection{Datasets and Baselines}
We evaluate our approach on the Dragon Lake Parking (DLP) dataset~\cite{shen2022parkpredict+}, which features dense, unstructured parking lots, and the Intersections Drone (inD) dataset~\cite{bock2020ind}, which comprises urban intersections with challenging vehicle–pedestrian interactions, complementing the parking scenarios in DLP. We follow the dataset split of ParkDiffusion~\cite{wei2025parkdiffusion} for both datasets. For DLP, we use 51750 samples for training and 5750 samples for validation. For inD, we choose “Bendplatz” and “Frankenburg” scenarios, with 34014 samples for training and 4859 samples for validation.

To the best of our knowledge, there are no existing baselines that jointly learn ego intention prediction and conditional joint trajectory prediction.
We therefore compare to strong ego-conditioned predictors that accept externally supplied ego inputs: WIMP~\cite{khandelwal2020wimp}, SceneTransformer~\cite{ngiam2022scenetransformer}, ScePT~\cite{chen2022scept}, MotionLM~\cite{seff2023motionlm}, and DTPP~\cite{huang2024dtpp}.
Since our implementation builds on ParkDiffusion~\cite{wei2025parkdiffusion}, which yields per-agent multimodal marginal predictions, we include a naive random joint baseline that assembles $K_\text{scene}$ joint scenes by uniformly sampling one marginal per agent. We denote this adapted baseline ParkDiffusion$^\dag$ and report its performance.
For comparability in parking layouts, we use a shared vectorized‑map frontend taken from ParkDiffusion for all methods and keep each baseline’s core architecture intact. This avoids layout‑specific feature biases while preserving each method’s modeling capacity.

\subsubsection{Evaluation Metrics}
We predict \(K_{\text{scene}}\) joint scenes per data sample. 
Oracle (min–) metrics take the best over the \(K_{\text{scene}}\) candidates. The 
{final (f–)} metrics evaluate the top‑$1$ scene selected by the model. 
ADE/FDE are average/final displacement errors in meters. 
MR counts a miss if none of the \(K_{\text{scene}}\) candidates have all valid agents with FDE \(\le \SI{2.0}{\meter}\) (scene‑level miss rate). 
OR denotes the overlap (collision) rate on the evaluated scene, where a scene is flagged if any vehicle-vehicle, vehicle-pedestrian, or vehicle-obstacle contact occurs. This extent of OR beyond agent-agent contacts to include agent-obstacle contacts is a special design feature from us for parking environments, where collisions between vehicles and obstacles frequently occur.
mAP is the area under the precision–recall curve over the \(K_{\text{scene}}\) joint candidates per sample, labeling a candidate positive if all valid agents have FDE \(\le \SI{2.0}{\meter}\). \looseness=-1

\subsubsection{Implementation Details}
All positions are in an ego‑centric frame, pre‑aligned by preprocessing. 
We observe \(T_p{=}10\) past steps and predict \(T_f{=}10\) future steps at \SI{0.4}{\second} intervals for a horizon of \SI{4.0}{\second}.  
Stage~1 trains the tokenizer for 20 epochs and proposes \(K_{\text{intent}}{=}6\) tokens.  
Stage~2 freezes the tokenizer and trains the conditional predictor for 20 epochs, decoding \(M{=}6\) marginals per agent and assembling \(K_{\text{scene}}{=}6\) joint candidates.  
Counterfactual Knowledge Distillation (CKD) is applied with probability \(p_{\mathrm{cf}}{=}0.5\) per batch. The teacher path uses the EMA decoder under a sampled non‑GT token and refines that prediction with \(t_{\text{steps}}{=}5\) denoiser steps. 
These parameters remain the same in benchmarking experiments, and in ablation studies, we state otherwise.
We use AdamW for training with a learning rate of \(1\mathrm{e}{-4}\) on a cosine schedule. We train the model on two NVIDIA A100 GPUs. \looseness=-1

\subsection{Quantitative Analysis}
\label{subsec:baseline}
On the DLP validation split, \method achieves the best oracle metrics among the compared methods.
For final metrics, \method yields the best f‑FDE, f‑MR, f‑OR, and f-mAP, with f-ADE of \qty{0.36}{\meter} following the best of \qty{0.34}{\meter} by MotionLM.
On the inD validation split, \method achieves the best oracle metrics in minFDE, minMR, minOR, mAP, and the second-best minADE of \qty{0.42}{\meter}.
For final metrics, we achieve the best in f-FDE, f-MR, f-OR, f-mAP. DTPP model achieves the best f-ADE of \qty{0.61}{\meter}.
We observe consistent trends across datasets. In both datasets, we achieve the best FDE, MR, OR, mAP for both oracle and final. However, minADE and f-ADE are less satisfying because the knowledge learned from counterfactual knowledge distillation and the safety-guided denoiser emphasize safety instead of efficiency. In real-world driving, human drivers sometimes adopt a more aggressive strategy for efficiency. 

\begin{table}
\centering
\caption{Main Components Ablation}
\vspace*{-0.3cm}
\label{tab:abl_components}
\setlength\tabcolsep{5pt}
\begin{threeparttable}
\begin{tabular}{l|ccccc}
\toprule
\textbf{Method} & \textbf{minADE} & \textbf{minFDE} & \textbf{MR} & \textbf{OR} & \textbf{mAP} \\
\midrule
ParkDiffusion$^\dag$~\cite{wei2025parkdiffusion} & 0.52 & 1.06 & 0.48 & 0.21 & 0.68 \\
+ JS      & 0.37 & 0.75 & 0.38 & 0.12 & 0.85 \\
+ JS + SGD & 0.32 & 0.60 & 0.26 & 0.05 & 0.92 \\
+ JS + SGD + CKD  & \textbf{0.29} & \textbf{0.56} & \textbf{0.23} & \textbf{0.03} & \textbf{0.97} \\
\bottomrule
\end{tabular}
Ablation on the main components. 
Starting from the vanilla ParkDiffusion$^\dag$, we add in sequence Joint Selector (JS), Safety-Guided Denoiser (SGD), and CKD to quantify each contribution.
Evaluation metrics are the oracle ones including minADE, minFDE, MR, OR, and mAP.
\end{threeparttable}
\vspace{-0.2cm}
\end{table}

\begin{table}
\centering
\caption{Ego Intention Tokenizer Ablation}
\vspace*{-0.3cm}
\label{tab:abl_tokenizer}
\setlength\tabcolsep{5pt}
\begin{threeparttable}
\begin{tabular}{l|ccccc}
\toprule
\textbf{Setting} & \textbf{minADE} & \textbf{minFDE} & \textbf{MR} & \textbf{OR} & \textbf{mAP} \\
\midrule
\multicolumn{6}{l}{\emph{Capacity $K_{\text{intent}}$}} \\
\cmidrule(lr){1-6}
\quad 1 (GT token)   & 0.31 & 0.60 & 0.27 & 0.04 & 0.96 \\
\quad 6               & \textbf{0.29} & \textbf{0.56} & \textbf{0.23} & \textbf{0.03} & \textbf{0.97} \\
\quad 12              & 0.30 & 0.58 & \textbf{0.23} & \textbf{0.03} & \textbf{0.97} \\
\quad 18              & 0.34 & 0.62 & 0.25 & 0.04 & 0.94 \\
\midrule
\multicolumn{6}{l}{\emph{Token source at $K_{\text{intent}}{=}6$}} \\
\cmidrule(lr){1-6}
\quad Stage~1 ranking  & \textbf{0.29} & \textbf{0.56} & \textbf{0.23} & \textbf{0.03} & \textbf{0.97} \\
\quad Stage~1 random      & 0.33 & 0.63 & 0.34 & 0.07 & 0.92 \\
\quad GT + noise    & 0.31 & 0.60 & 0.27 & 0.04 & 0.96 \\
\bottomrule
\end{tabular}
Ablation on the Ego Intention Tokenizer. Intention capacity describes how many intention tokens we generate for the downstream CKD module. Token source includes three token selection strategies, namely ranking Stage~1 outputs and select the highest possible ones (Stage~1 ranking), random sampling from the Stage~1 outputs (Stage~1 random), and sampling around ground truth token with random noise (GT + noise).
\end{threeparttable}
\vspace{-0.3cm}
\end{table}

\subsection{Qualitative Analysis}
\cref{fig:main_figure} visualizes the prediction results of \method on both datasets. 
On the DLP dataset, our model accurately predicts the ego intention, and the resulting conditional joint trajectory predictions closely align with the ground truth. 
When conditioned on counterfactual ego intentions, other agents exhibit plausible reactive behaviors to avoid potential collisions.
For instance, in the second panel of the top row, a vehicle positioned to the ego's front-left adjusts its trajectory to yield, thereby preventing a potential conflict.
Similarly, on the inD dataset, the model effectively captures the scene's dynamic patterns when conditioned on the most likely intention.
In the third panel, a pedestrian is predicted to yield in response to the ego vehicle's potential acceleration. 

We find that our model presents a conservative tendency in the predictions across both datasets. 
For example, on the DLP dataset, the vehicle yielding to the ego agent we mentioned actually yields further left to create more space from the ego vehicle, even though this brings it closer to more vulnerable pedestrian agents.
This behavior may be attributed to a lack of prioritized collision penalties in the model's training objective.
A similar conservative trend is observed on the inD dataset, where a pedestrian behind the ego vehicle is predicted to move slower than its ground truth, and two other pedestrians on the upper left adjust their paths to avoid a mutual collision.
Collectively, these qualitative examples demonstrate the effectiveness of the conditional joint prediction model with safety constraints, matching the strong quantitative results where our method achieved the lowest oracle and final OR over all baselines.

\subsection{Ablation Studies}
\label{subsec:ablations}

We perform ablation studies on the DLP validation dataset to provide insights into the impact of the design of individual components. 
\subsubsection{Components Analysis}
We ablate on different components and present results in \cref{tab:abl_components}. Random composition of independent marginals (ParkDiffusion$^\dag$) typically yields poor MR/OR and weak mAP. Adding the Joint Selector (JS) converts it to a truly learned joint prediction, improving all the metrics by a large margin. The Safety-Guided Denoiser (SGD) further improves the overall performance, and most notably, the safety-critical OR metrics. The Counterfactual Knowledge Distillation (CKD) transfers teacher‑refined behavior to the student, continues to improve the general performance across all dimensions.

\begin{table}
\centering
\caption{Counterfactual Knowledge Distillation Ablation}
\vspace*{-0.3cm}
\label{tab:abl_ckd}
\setlength\tabcolsep{5pt}
\begin{threeparttable}
\begin{tabular}{l|ccccc}
\toprule
\textbf{Setting} & \textbf{minADE} & \textbf{minFDE} & \textbf{MR} & \textbf{OR} & \textbf{mAP} \\
\midrule
\multicolumn{6}{l}{\emph{Counterfactual sampling probability $p_{\mathrm{cf}}$}} \\
\cmidrule(lr){1-6}
\quad 0.00 & 0.32 & 0.60 & 0.26 & 0.05 & 0.92 \\
\quad 0.25 & 0.31 & 0.59 & 0.28 & 0.04 & 0.96 \\
\quad 0.50 & \textbf{0.29} & \textbf{0.56} & \textbf{0.23} & \textbf{0.03} & \textbf{0.97} \\
\quad 0.75 & 0.35 & 0.64 & 0.27 & 0.04 & 0.94 \\
\midrule
\multicolumn{6}{l}{\emph{Teacher configuration}} \\
\cmidrule(lr){1-6}
\quad None         & 0.32 & 0.60 & 0.26 & 0.05 & 0.92 \\
\quad EMA only     & 0.32 & 0.59 & 0.26 & 0.05 & 0.93 \\
\quad EMA + Denoiser(U)         & 0.30 & 0.58 & 0.25 & 0.05 & 0.93 \\
\quad EMA + SGD     & \textbf{0.29} & \textbf{0.56} & \textbf{0.23} & \textbf{0.03} & \textbf{0.97} \\
\bottomrule
\end{tabular}
Ablation on the Counterfactual Knowledge Distillation. We show the ablation from two perspectives: counterfactual sampling probability describing the weight of counterfactual learning over supervised learning, and teacher configurations including None (no CKD at all), EMA only, EMA + Denoiser(U) (EMA + Denoiser without safety guided potential functions), and EMA + SGD (EMA + Safety-Guided Denoiser).
\end{threeparttable}
\vspace{-0.3cm}
\end{table}

\subsubsection{Ego Intention Tokenizer}
We ablate the ego intention tokenizer capacity and token source in \cref{tab:abl_tokenizer}. We observe that the performance saturates around \(K_{\text{intent}}{=}6\). Increasing the number of counterfactual tokens to 12 brings negative effects to accuracy, and when it is increased to 18, performance across all dimensions drops by a large margin, indicating that the counterfactual knowledge starts to confuse the main joint predictor.
For the token selection strategy in CKD training, the Stage~1 ranking strategy outperforms the other two, reflecting that this strategy brings extra environment awareness knowledge to Stage~2 to help the learning process.

\subsubsection{Counterfactual Knowledge Distillation (CKD)}
We ablate on the counterfactual knowledge distillation from two perspectives, counterfactual sampling probability and teacher configuration. We present the results in \cref{tab:abl_ckd}. We find that balancing the supervised learning with the CKD modules achieves the best performance. When CKD module has larger weights, the performance drops quickly. Regarding different teacher configurations, adding EMA module alone helps stabilize the training but leads to no explicit gain in the metrics. Adding a denoiser without safety guidance based on EMA helps to improve all metrics slightly. With the Safety-Guided Denoiser (SGD) added in the last step, the MR, OR, mAP scores improve by a large margin.

%% file: sections/5_conclusion.tex
\section{Conclusion} 
We presented \method, a two-stage approach for ego intention conditioned joint trajectory prediction in automated parking. 
Stage~1 predicts a compact set of ego intention tokens, and Stage~2 performs intention conditioned joint prediction augmented by counterfactual knowledge distillation and safety-guided modules.
Experiments show that \method achieves the state-of-the-art quantitative results, and present qualitative what-if visualizations.
Limitations include reliance on handcrafted geometric potentials in the denoiser and the remaining gap between offline counterfactual supervision and closed‑loop control. 
Future work could explore how to learn guidance potentials from data and how to couple the conditional predictor with smarter decision making, planning, and control.